%% file: egpaper_for_review.tex
\documentclass[10pt,twocolumn,letterpaper]{article}

\usepackage{cvpr}
\usepackage{times}
\usepackage{epsfig}
\usepackage{graphicx}
\usepackage{amsmath}
\usepackage{amssymb}

\usepackage{subcaption}
\usepackage{bbding}
\usepackage{pifont}
\usepackage[normalem]{ulem}
\usepackage{multirow}
\usepackage[table,xcdraw]{xcolor}

\newcommand{\beas}{\begin{eqnarray*}}
\newcommand{\eeas}{\end{eqnarray*}}
\newcommand{\bea}{\begin{eqnarray}}
\newcommand{\eea}{\end{eqnarray}}
\newcommand{\bes}{\begin{equation*}}
\newcommand{\ees}{\end{equation*}}
\newcommand{\be}{\begin{equation}}
\newcommand{\ee}{\end{equation}}
\newcommand{\cL}{{\cal L}}
\newcommand{\KL}{\operatorname{KL}}
\newcommand{\sE}{\mathbb{E}}
\newcommand{\cN}{{\cal N}}

\makeatletter
\usepackage{xspace}
\def\@onedot{\ifx\@let@token.\else.\null\fi\xspace}
\DeclareRobustCommand\onedot{\futurelet\@let@token\@onedot}

\newcommand{\figref}[1]{Fig\onedot~\ref{#1}}
\newcommand{\equref}[1]{Eq\onedot~\eqref{#1}}
\newcommand{\secref}[1]{Sec\onedot~\ref{#1}}
\newcommand{\tabref}[1]{Tab\onedot~\ref{#1}}

\def\eg{\emph{e.g}\onedot} 
\def\ie{\emph{i.e}\onedot} 
 
\def\etc{\emph{etc}\onedot} 
\def\wrt{w.r.t\onedot} 
\def\etal{\emph{et al}\onedot}

\makeatother

\usepackage[pagebackref=true,breaklinks=true,letterpaper=true,colorlinks,bookmarks=false]{hyperref}

\cvprfinalcopy % *** Uncomment this line for the final submission

\def\cvprPaperID{2922} % *** Enter the CVPR Paper ID here

\ifcvprfinal\fi
\begin{document}

%%%%%%%%% TITLE
\title{Creativity: Generating Diverse Questions using Variational Autoencoders}

\author{Unnat Jain\thanks{~indicates equal contributions.}\\
UIUC\\
{\tt\small uj2@illinois.edu}
\and
Ziyu Zhang\footnotemark[1]\\
Northwestern University\\
{\tt\small zzhang@u.northwestern.edu}
\and
Alexander Schwing\\
UIUC\\
{\tt\small aschwing@illinois.edu}
}

\maketitle
\thispagestyle{empty}

\input{abs}
\input{intro}
\input{rel}
\input{app}
\input{exp}

\input{conc}

\noindent\textbf{Acknowledgements:} We thank NVIDIA for providing the GPUs used for this research.

% \clearpage

{\small
\bibliographystyle{ieee}
\bibliography{alex}
}

% \clearpage

% \input{supp}

\end{document}

%% file: abs.tex
\begin{abstract}
Generating diverse questions for given images is an important task for computational education, entertainment and AI assistants. Different from many conventional prediction techniques is the need for algorithms to generate a diverse set of plausible questions, which we refer to as ``creativity''. In this paper we propose a  creative algorithm for visual question generation which combines the advantages of variational autoencoders with long short-term memory networks. We demonstrate that our framework is able to generate a large set of varying questions given a single input image.
%``Visual Creativity'' is a term referring to techniques which are designed to create new worlds of visual ideas. In this paper, we propose to use a truly creative mechanism to generate questions given images.
\end{abstract}

%% file: intro.tex
%!TEX root = egpaper_for_review.tex
\section{Introduction}
%What is the problem and why is it important
Creativity is  a mindset that can be cultivated, and ingenuity is not a gift but a skill that can be  trained. Pushing this line of thought one step further, in this paper, we ask: why are machines and in particular computers not creative? Is it due to the fact that our environment is oftentimes perfectly predictable at the large scale? Is it because the patterns of our life provide convenient shortcuts, and creativity is not a necessity to excel in such an environment?

Replicating human traits on machines  is a long-standing goal, and remarkable recent steps to effectively extract representations from data~\cite{BengioARXIV2013,LeCunNature2015} have closed the gap between human-level performance and `computer-level' accuracy on a large variety of tasks such as object classification~\cite{KrizhevskyNIPS2012}, %\unnat{needs correct reference (~\cite{KrizhevskyNIPS2012} is classification)}, 
speech-based translation~\cite{SutskeverNIPS2014}, and language-modeling~\cite{HintonSPM2012}. There seems no need for computers to be creative. 
However, creativity is crucial if existing knowledge structures fail to yield the desired outcome. %Humans switch effortlessly between invention and repetition. While computers are good at the latter, they fail miserably at the former probably more important task. 
We cannot hope  to encode all logical rules into algorithms, or all observations into features, or  all data into representations. Hence, we need novel frameworks to automatically mine and implicitly characterize knowledge databases, and we need algorithms which are creative at combining those database entries.

%How have people solved the problem so far
Generative modeling tools can be used for those tasks since they aim at characterizing the distribution from which  datapoints are sampled. 
Classical generative models such as restricted Boltzmann machines~\cite{HintonScience2006}, probabilistic semantic indexing~\cite{Hofmann1999} or latent Dirichlet allocation~\cite{Blei2006} sample from complex distributions. 
Instead, in recent years, significant progress in  generative modeling suggests to sample from simple distributions and to subsequently transform the sample via function approximators to yield the desired output.  Variational autoencoders~\cite{KingmaARXIV2013,KingmaNIPS2014} and adversarial nets~\cite{GoodfellowARXIV2014} are among   algorithms which follow this paradigm.  %In contrast to classical methods, adversarial nets and variational autoencoders sample from simple distributions and subsequently transform the sample via function approximators to yield the desired output. 
Both, variational autoencoders and adversarial nets have successfully been applied to a variety of tasks such as image generation, sentence generation \etc.

\begin{figure}[t]
\centering
\includegraphics[width=\linewidth]{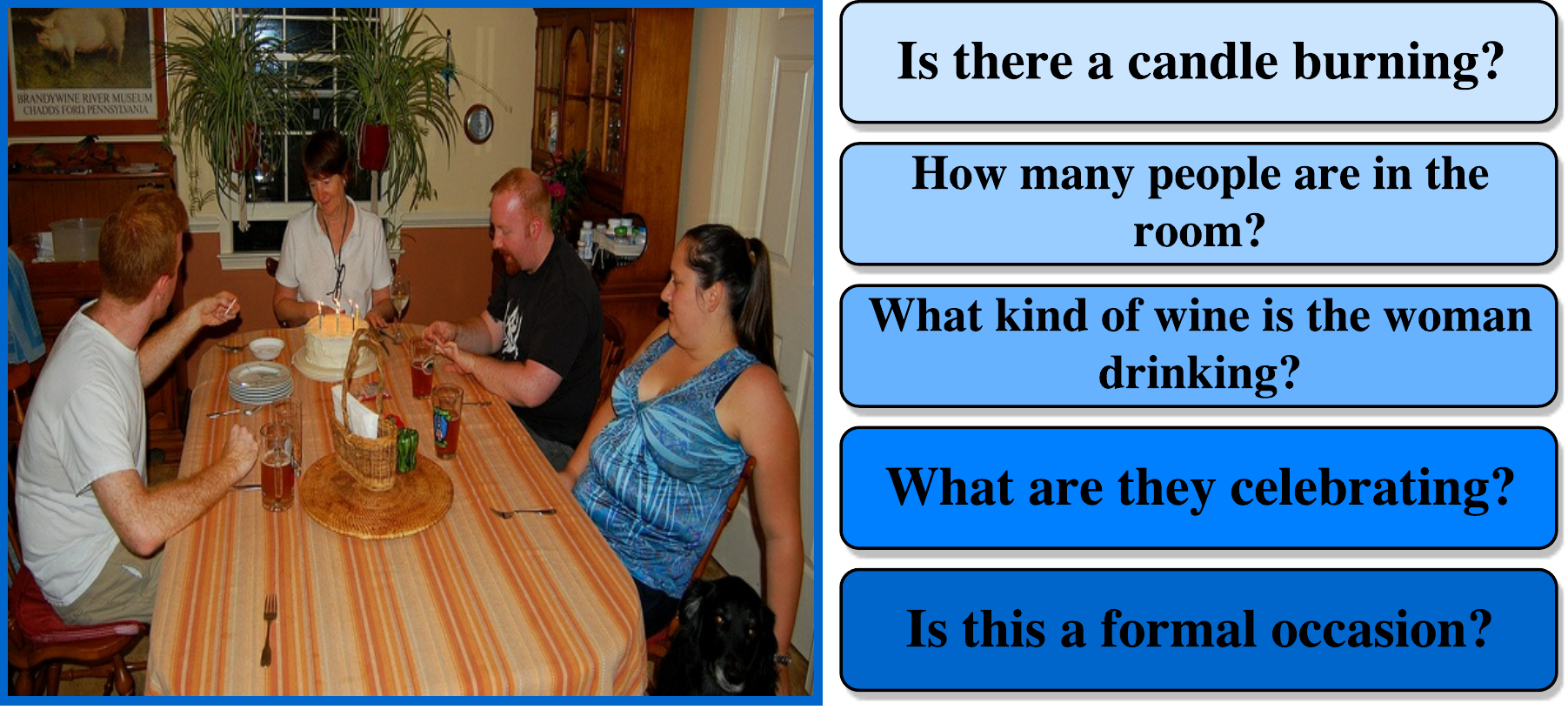}
\vspace{-0.5cm}
\caption{Questions generated by our approach - \textit{literal} to \textit{inferential}, \ie, questions following visual content and  questions requiring scene understanding and prior information about  objects. %This image shows a gradient from literal to inferential questions our system produced for the image.
}
\label{fig:Teaser}
\vspace{-0.5cm}
\end{figure}
\begin{figure*}
\vspace{-0.0cm}
\centering
\includegraphics[width=\linewidth]{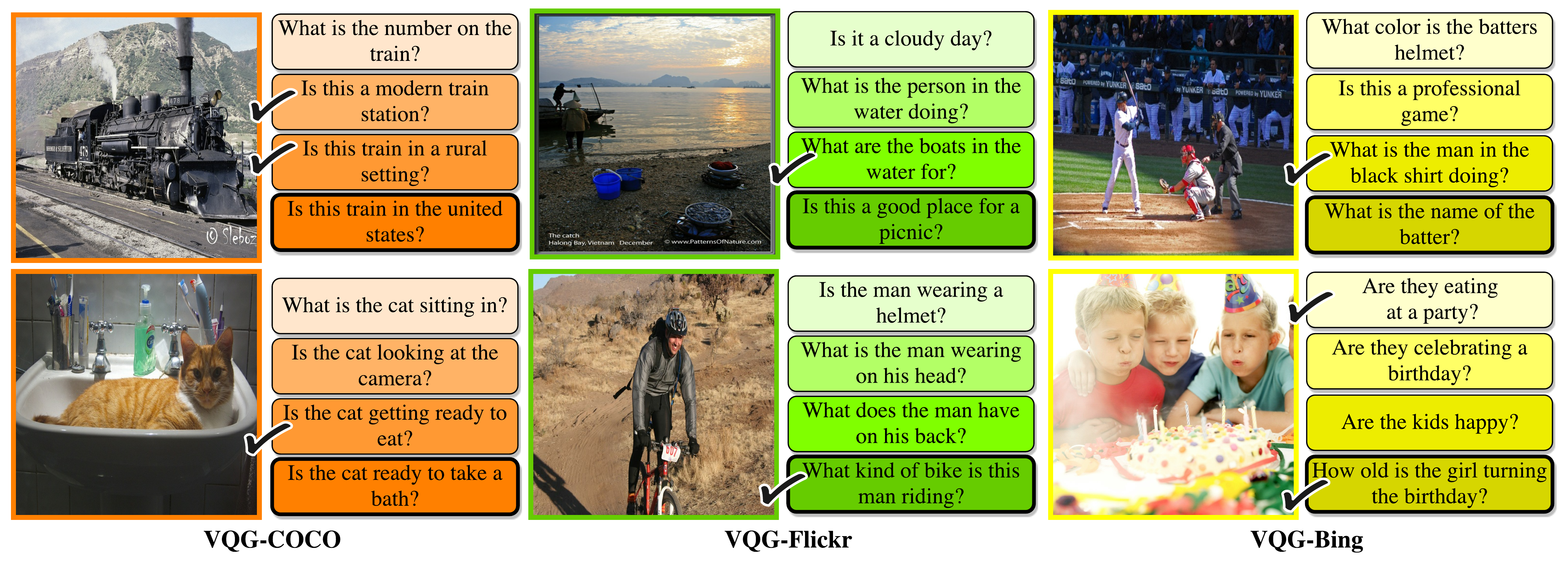}
\vspace{-0.9cm}
\caption{Examples of questions generated by our VQG algorithm. 
Darker colored boxes contain questions which are more inferential.  
%The color gradient signifies the need for more visual understanding and prior knowledge to answer the question.
%These generated 
Our questions include queries about numbers and scanty clouds  showing its visual recognition strength. Questions on events, type of sport and motion demonstrate an ability to understand scenes and actions. Unlike questions on colors, counts and shapes, the questions in $\boxed{\textbf{bold box}}$ are exemplars of how diverse our model is. It fuses visual information with context to ask questions which cannot be answered simply by looking at the image. Its answer requires prior (human-like) understanding of the objects or scene. The questions with \textbf{bold ticks} (\ding{52}) are  generated by our VQG model which never occurred during training (what we refer to as `unseen' questions). }
\label{fig:qual}
\vspace{-0.2cm}
\end{figure*}

%How do we propose to solve the problem
In this work we will use those algorithms for the novel task of visual question generation as opposed to visual question answering. Visual question generation is useful in a variety of areas where it is important to engage the user, \eg, in computational education, for AI assistants, or for entertainment. Retaining continued interest and curiosity is crucial in all those domains, and can only be achieved if the developed system is continuously exposing novel aspects rather than repeating a set of handcrafted traits. 
Visual question generation closes the loop to question-answering and diverse questions enable engaging conversation, helping AI systems such as driving assistants, chatbots, \etc, to perform better on Turing tests. 
Concretely, consider a program that aims at teaching kids to describe images. Even if 100 exciting questions are provided, the program will eventually exhaust all of them and we quickly put the program aside mocking about repetitive `behavior.' To alleviate this issue, we argue that creative mechanisms are important, particularly in domains such as question generation.

%\alex{cite work by dhruv on diversity} \unnat{~\cite{VijayakumarARXIV2016}}

% 
In this paper we propose a technique for generating diverse questions that is based on generative models. More concretely, we follow the variational autoencoder paradigm rather than adversarial nets, because training seems oftentimes more stable. % \unnat{reviewer requested a reference for this claim}. 
We learn to embed a given question and the features from a corresponding image into a low-dimensional latent space. During inference, \ie, when we are given a new image, we generate a question by sampling from the latent space and subsequently decode the sample together with the feature embedding of the image to obtain a novel question. We illustrate some images and a subset of the generated questions in \figref{fig:Teaser} and \figref{fig:qual}. Note the diversity of the generated questions some of which are more literal while others are more inferential.

%What do we evaluate it on
In this paper we evaluate our approach on the VQG - COCO, Flickr and Bing datasets~\cite{VQG}. We demonstrate that the proposed technique is able to ask a series of remarkably diverse questions given only an image as input.

%% file: rel.tex
%!TEX root = egpaper_for_review.tex

\section{Related Work}
When considering generation of text from images, caption and paragraph generation~\cite{JohnsonCVPR2016,BarnardJMLR2003,ChenCVPR2015,DonahueARXIV2014,FangCVPR2015,FarhadiECCV2010,ChoARXIV2015,KarpathyCVPR2015,KirosTACL2015,KulkarniCVPR2011,MaoARXIV2014,SocherTACL2014,VinyalsCVPR2015,XuICML2015}, as well as visual question answering~\cite{AnatolICCV2015,GaoNIPS2015,MalinowskiICCV2015,RenNIPS2015,ShihCVPR2016,XiongICML2016,XuARXIV2015,YangCVPR2016,FukuiARXIV2016,KimARXIV2016,ZitnickAI2016,AndreasCVPR2016,DasARXIV2016,JabriARXIV2016,YuARXIV2015,ZhouARXIV2015,WuARXIV2016,MaARXIV2015,LuARXIV2016,JabriARXIV2016} come to mind immediately. We first review those tasks before discussing work related to visual question generation and generative modeling in greater detail.

\noindent{\bf Visual Question Answering and Captioning} are two tasks that have received a considerable amount of attention in recent years. Both assume an input image to be available during inference. For visual question answering we also assume a question to be provided. For both tasks a variety of different models have been proposed and %, %even though the difference between attention based mechanisms and carefully designed neural net baselines is small, 
attention mechanisms have emerged as a valuable tool because they permit to catch a glimpse on what the generally hardly interpretable neural net model is concerned about.

\noindent{\bf Visual Question Generation}  is a  task that has been proposed very recently and is still very much an open-ended topic. Ren \etal~\cite{RenNIPS2015} proposed a rule-based algorithm to convert a given sentence into a corresponding question that has a single word answer. %Utilizing the generative model for captions proposed by Vinyals \etal~\cite{VinyalsCVPR2015}, their algorithm converted the caption to a question.
Mostafazadeh \etal~\cite{VQG} were the first to learn a question generation model using human-authored questions instead of machine-generated captions. They focus on creating a `natural and engaging' question. Recently, Vijayakumar \etal~\cite{VijayakumarARXIV2016} have shown preliminary results for this task as well.

We think that visual question generation is an important task for two reasons. First, the task is dual to visual question answering and by addressing both tasks we can close the loop. Second, we think the task is in spirit similar to `future prediction' in that a reasonable amount of creativity has to be encoded in the model. Particularly the latter is rarely addressed in the current literature. For example, Mostafazadeh \etal~\cite{VQG} obtain best results by generating a single question per image using a forward pass of image features through a layer of LSTMs or gated recurrent units (GRUs). Vijayakumar \etal~\cite{VijayakumarARXIV2016} show early results of question generation by following the same image caption generative model~\cite{VinyalsCVPR2015} as COCO-QA, but by adding a diverse beam search step to boost diversity.

Both techniques yield encouraging results. However in~\cite{VQG} only a single question is generated per image, while the approach discussed in~\cite{VijayakumarARXIV2016} generates diverse questions by sampling from a complicated energy landscape, which is intractable in general~\cite{GimpelEMNLP2013,BatraECCV2012}. In contrast, in this paper, we follow more recent generative modeling paradigms by sampling form a distribution in an encoding space. The encodings are subsequently  mapped to a high-dimensional representation  using, in our case, LSTM nets, which we then use to generate the question.

\noindent{\bf Generative Modeling}
 of data is a longstanding goal. First attempts such as   k-means clustering~\cite{Lloyd1982} and the Gaussian mixture models~\cite{Dempster1977}  restrict the class of considered distributions severely, which leads to significant modeling errors when considering  complex distributions required to model objects such as sentences.
Hidden Markov models~\cite{Baum1966}, probabilistic latent semantic indexing~\cite{Hofmann1999}, latent Dirichlet allocation~\cite{BleiJMLR2003} and restricted Boltzmann machines~\cite{Smolensky1986,HintonScience2006} extend the classical techniques. Those extensions work well when carefully tuned to specific tasks but  struggle to model the high ambiguity inherently tied to images.

More recently deep nets have been used as function approximators for generative modeling, and, similar to deep net performance in many other areas, they produced extremely encouraging results~\cite{GoodfellowARXIV2014,DentonNIPS2015,RadfordARXIV2016}. Two very successful approaches  are referred to as generative adversarial networks (GANs)~\cite{GoodfellowARXIV2014} and variational auto-encoders (VAEs)~\cite{KingmaARXIV2013}. However their success  relies on a variety of tricks for successful training~\cite{SalimansARXIV2016,GoodfellowARXIV2014,RadfordARXIV2016,BurdaARXIV2015}.

Variational auto-encoders (VAEs) were first introduced by Kingma and Welling~\cite{KingmaARXIV2013} and they were quickly adopted across different areas. They were further shown to be useful in the semi-supervised setting~\cite{KingmaNIPS2014}.
Conditional VAEs were recently considered by Yan \etal~\cite{YanARXIV2015}. Moreover, it was also shown by Krishnan \etal~\cite{KrishnanARXIV2015} and Archer \etal~\cite{ArcherARXIV2015} how to combine VAEs with continuous state-space models. In addition, Gregor \etal~\cite{GregorARXIV2015} and Chung \etal~\cite{ChungNIPS2015} demonstrated how to extend VAEs to sequential modeling, where they focus on RNNs.
%Work by Johnson \etal~\cite{JohnsonARXIV2016} propose a general framework for exponential family densities. While Johnson \etal~\cite{JohnsonARXIV2016} consider structure in the latent space and Sohn \etal~\cite{SohnNIPS2015} focus on sampling based methods. 

%While many of those extensions are applicable in the case of visual question generation, we leave those extensions to future work and first focus on carefully assessing the approach described in the following.

%% file: app.tex
%!TEX root = egpaper_for_review.tex
\section{Approach}
For the task of visual question generation, demonstrated in \figref{fig:qual}, we rely on variational autoencoders (VAEs). Therefore, in the following, we first provide background on VAEs before  presenting the proposed approach.

\begin{figure}[t]
\centering
\includegraphics[width=\linewidth]{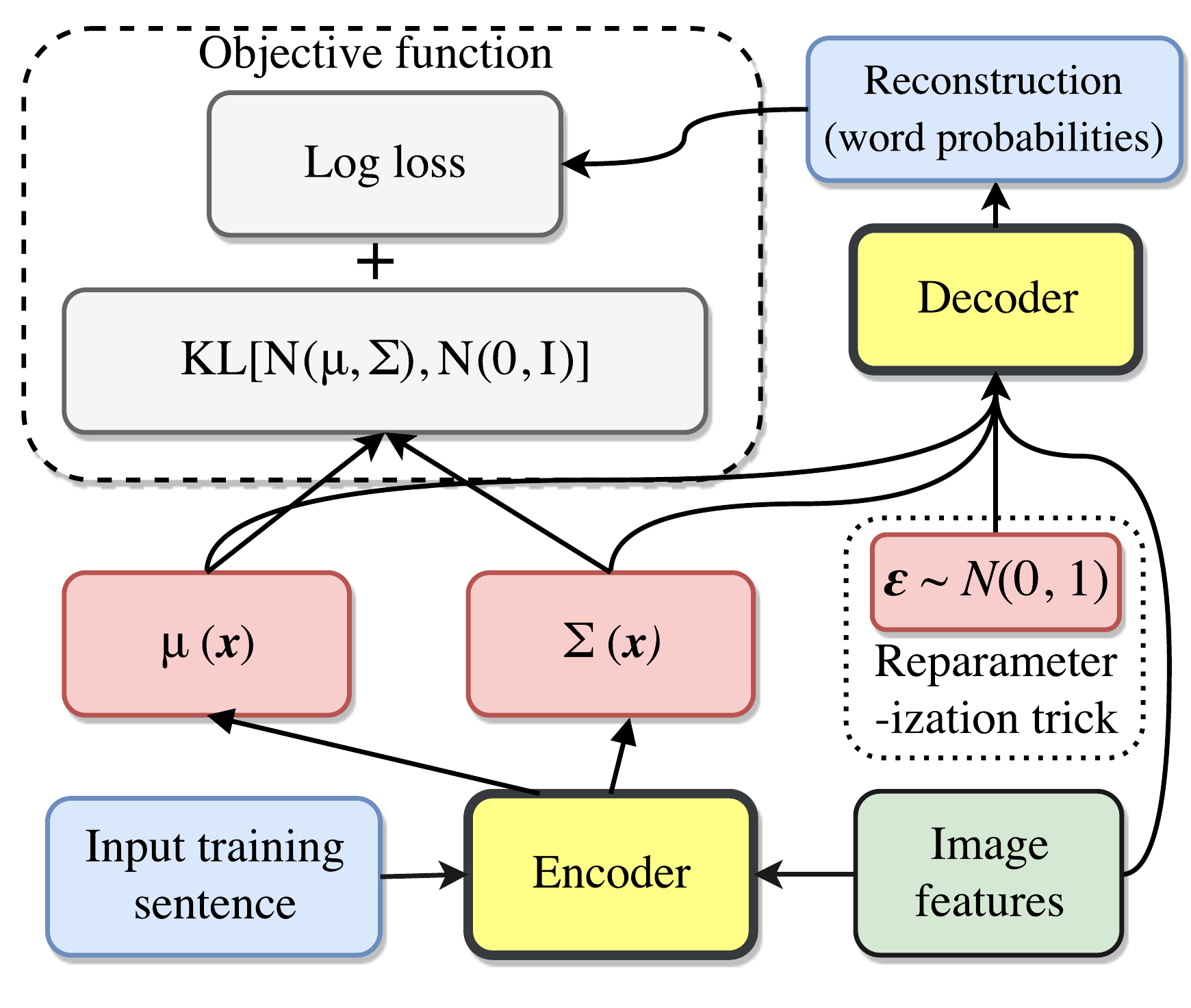}
\vspace{-0.9cm}
\caption{High level VAE overview of our approach.}
\label{fig:HighLevel}
\vspace{-0.3cm}
\end{figure}

\subsection{Background on Variational Autoencoders}
Following common techniques for latent variable models, VAEs assume that it is easier to optimize a parametric distribution $p_\theta(x,z)$ %$p_\theta(y|x,z)$, 
defined over both the variables $x$, in our case the words of a sentence, as well as a latent representation $z$. % $\cZ$. %, \ie, a distribution . % that is flexible enough.
By introducing a data-conditional latent distribution $q_\phi(z|x)$ the log-likelihood of a datapoint $x$, \ie, $\ln p_\theta(x)$, can be re-written as follows:
\bea
\ln p_\theta(x) &\hspace{-0.2cm}=&\hspace{-0.2cm} \sum_z q_\phi(z|x)\ln p_\theta(x) %= \sum_z q_\phi(z|x)\ln\frac{p_\theta(x,z)q_\phi(z|x)}{p_\theta(z|x)q_\phi(z|x)}
\nonumber\\
&\hspace{-2.0cm}=&\hspace{-1.0cm} \sum_z \left[q_\phi(z|x)\ln\frac{p_\theta(x,z)}{q_\phi(z|x)} - q_\phi(z|x)\ln\frac{p_\theta(z|x)}{q_\phi(z|x)}\right]\nonumber\\
&\hspace{-2.0cm}=&\hspace{-1.0cm}  \cL(q_\phi(z|x),p_\theta(x,z)) + \KL(q_\phi(z|x),p_\theta(z|x)).\label{eq:LogLik1}
\eea

Since the KL-divergence is non-negative, $\cL$ is a lower bound on the log-likelihood $\ln p_\theta(x)$.
Note that computation of the KL-divergence is not possible because of the unknown and generally intractable posterior $p_\theta(z|x)$.
However when choosing a parametric distribution $q_\phi(z|x)$ with  capacity large enough to fit the posterior $p_\theta(z|x)$,  the log-likelihood \wrt $\theta$ is optimized by instead maximizing the lower bound \wrt both $\theta$, and $\phi$. Note that the maximization of $\cL$ \wrt $\phi$ reduces the difference between the lower bound $\cL$ and the log-likelihood $\ln p_\theta(x)$.
%We further note that the log likelihood, \ie, the left hand side of the aforementioned derivation, does not depend on the parameters $\phi$. We also assume the distribution $q_\phi(z|x)$ to have sufficiently large capacity such that we are able to match the intractable and unknown posterior $p_\theta(z|x)$.
%In this case, maximizing the log likelihood \wrt the parameters $\phi$ is equivalent to maximizing the lower bound $\cL(q_\phi(z|x),p_\theta(x,z))$. %, or alternatively, minimizing the KL-divergence, \wrt the parameters $\phi$. %, which is equivalent to minimizing the KL-divergence
%Since the KL divergence is non-negative, maximizing the log likelihood \wrt the parameters $\phi$ corresponds to choosing the parameters $\phi$ of the conditional $q_\phi(z|x)$ to be as close as possible to the data distribution $p_\theta(z|x)$. Note that we are not necessarily guaranteed to find parameters $\phi$ that exactly match $p_\theta(z|x)$, hence we cannot ignore the KL-divergence $\KL(q_\phi,p_\theta)$ when computing a derivative \wrt $\theta$.
%Investigating the result more carefully we note that maximizing the log-likelihood $\ln p_\theta(x)$ \wrt its parameters $\theta$ is equivalent to alternating maximizing of the lower bound $\cL$ \wrt the distributions $q_\phi$ and $p_\theta$.
Instead of directly maximizing the lower bound $\cL$ given in \equref{eq:LogLik1} \wrt $\theta,\phi$, dealing with a joint distribution $p_\theta(x,z)$ can be avoided via
\bea
\cL(q_\phi,p_\theta) &\hspace{-0.2cm}=&\hspace{-0.2cm} %\sum_z q_\phi(z|x)\ln\frac{p_\theta(x,z)}{q_\phi(z|x)} = 
\sum_z q_\phi(z|x)\ln\frac{p_\theta(x|z)p_\theta(z)}{q_\phi(z|x)}\nonumber\\
&\hspace{-2.5cm}=&\hspace{-1.5cm} \sum_z q_\phi(z|x)\ln\frac{p_\theta(z)}{q_\phi(z|x)} + \sum_z q_\phi(z|x)\ln p_\theta(x|z)\nonumber\\
&\hspace{-2.5cm}=&\hspace{-1.5cm}-\KL(q_\phi(z|x),p_\theta(z)) + \sE_{q_\phi(z|x)}\left[\ln p_\theta(x|z)\right].\label{eq:LowerBound}
\eea
Note that $p_\theta(z)$ is a prior distribution over the latent space and $q_\phi(z|x)$ is modeling the intractable and unknown posterior $p_\theta(z|x)$. Intuitively the model distribution is used to guide the likelihood evaluation by focusing on highly probable regions.

%In a next step we combine the lower bound given in \equref{eq:LowerBound} with the log likelihood provided in \equref{eq:LogLik1}. To do so we first note that the KL-divergence terms are not exactly identical. It is however more than natural to assume that the latent object $z$ is independent of the observed data $x$, \ie, $p_\theta(z|x) = p_\theta(z)$. Hence, when additionally noting that maximization of the lower bound is equivalent to minimizing its negative value, we obtain the following program:
In a next step the expectation over the model distribution $q_\phi$ is approximated with $N$ samples $z_i\sim q_\phi$, \ie, after abbreviating  $\KL(q_\phi(z|x),p_\theta(z))$ with $\KL(q_\phi,p_\theta)$ we obtain:
\be
\min_{\phi,\theta} \KL(q_\phi,p_\theta)-\frac{1}{N}\sum_{i=1}^N \ln p_\theta(x|z^i), \hspace{0.1cm}\text{s.t.}\hspace{0.1cm} z^i \sim q_\phi.
\label{eq:ProgramSimple}
\ee

%Hereby we approximate the expectation over the distribution $q_\phi$ with $N$ samples $z_i\sim q_\phi$.

%The intuition behind this program is as follows.
%Given data $x$, \eg, an image, we want to encode it into a latent representation via $q_\phi(z|x,y)$. We expect this latent representation to be similar to a prior $p_\theta(z|x)$ which may or may not use some of the parameters in $\theta$. In addition 
%We aim at generating sample $z_i$ from a distribution $q_\phi$ which are well suited for reconstruction of the data sample $x$. %The latter is encouraged by minimizing the negative log likelihood $p_\theta(x|z^i)$.

In order to solve this program in an end-to-end manner, \ie, to optimize \wrt both the model parameters $\theta$ and the parameters $\phi$ which characterize the distribution over the latent space, it is required to differentiate through the sampling process.
To this end Kingma and Welling~\cite{KingmaARXIV2013} propose to make use of the `reparameterization trick.' For example, if we restrict $q_\phi(z|x)$ to be an independent Gaussian with mean $\mu_j$ and variance $\sigma_j$ for each component $z_j$ in $z = (z_1, \ldots, z_M)$, then we can sample easily via $z_j^i = \mu_j + \sigma_j \cdot \epsilon^i$ where $\epsilon^i \sim \cN(0,1)$. The means $\mu_j(x,\phi)$ and variances $\sigma_j(x,\phi)$ are parametric functions which are provided by the encoder. A general overview of VAEs is provided in \figref{fig:HighLevel}. %In this latter case we use the means and variances as the parameters of the model, \ie, $\phi = (\ldots, \mu_j, \ldots, \sigma_j, \ldots)$.

\begin{figure}[t]
\vspace{-0.5cm}
\centering
\includegraphics[width=\linewidth]{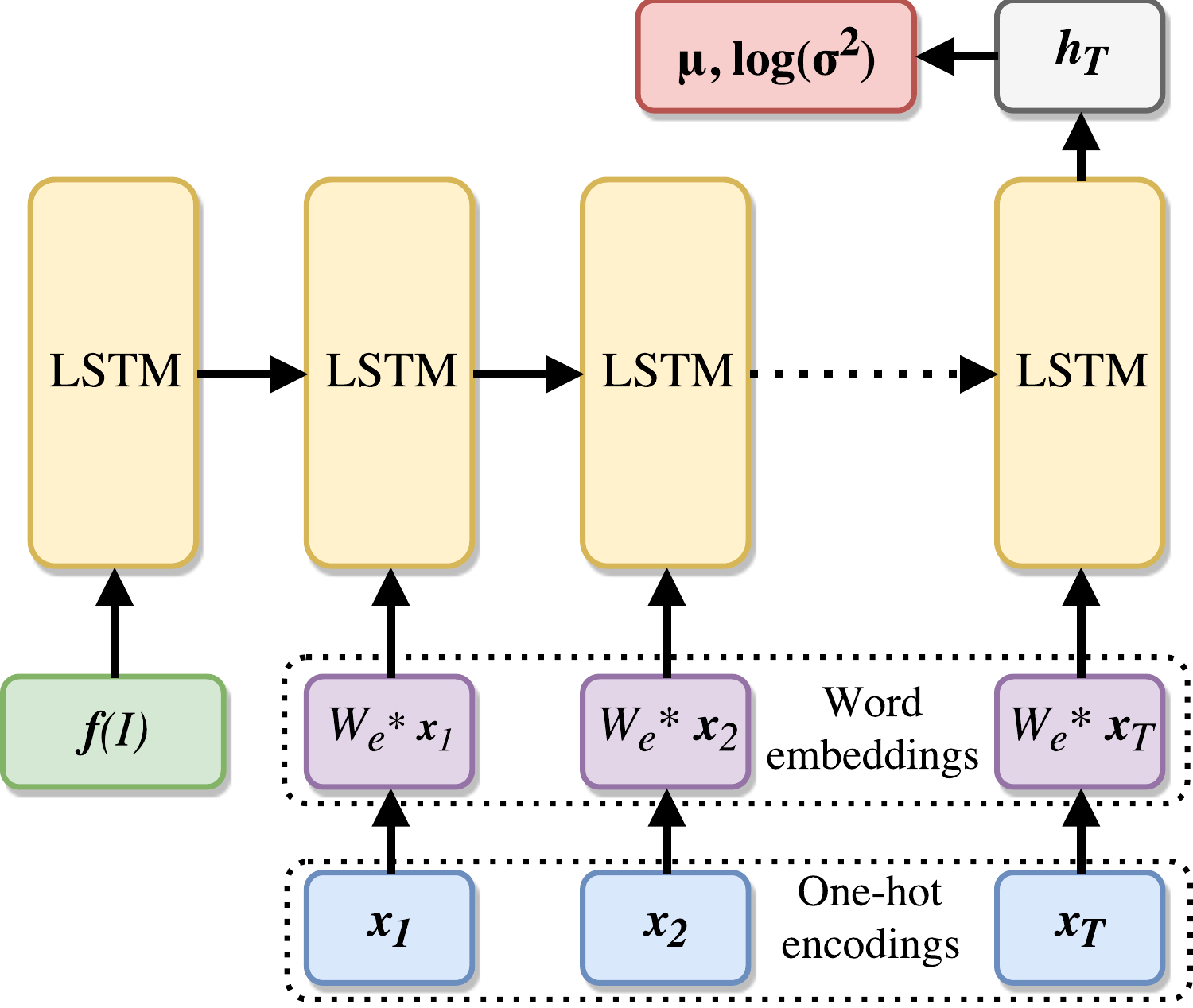}
\vspace{-0.7cm}
\caption{Q-distribution: The $V$-dimensional 1-hot encoding of the vocabulary (blue) gets embedded linearly via $W_e\in\mathbb{R}^{E\times V}$ (purple). Embedding and $F$-dimensional image feature (green) are the LSTM inputs, transformed to fit the $H$ dimensional hidden space. We transform the final hidden representation via two linear mappings to estimate mean and log-variance.}
\label{fig:QDist}
\vspace{-0.2cm}
\end{figure}

\begin{figure}[t]
\vspace{-0.5cm}
\centering
\includegraphics[width=\linewidth]{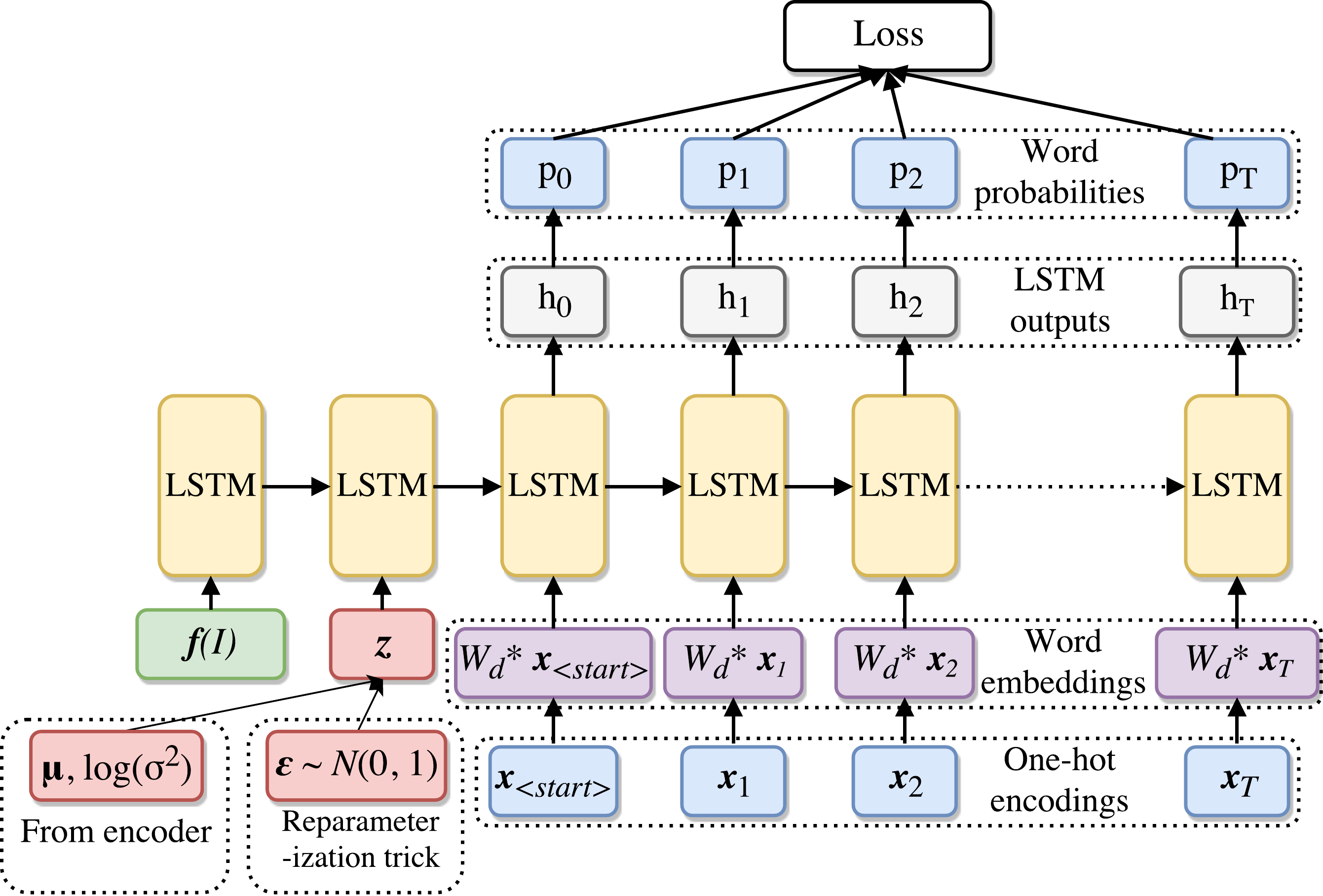}
\vspace{-0.7cm}
\caption{P-distribution: Input to the LSTM units are the $F$-dimensional image feature $f(I)$, the $M$-dimensional sample $z$ (transformed during training), and the $E$-dimensional word embeddings. To obtain a prediction we transform the $H$-dimensional latent space into the $V$-dimensional logits $p_i$.}
\label{fig:PDist}
\vspace{-0.2cm}
\end{figure}

\subsection{Visual Question Generation}
In the following we describe our technique for learning a high-dimensional embedding and for inference in greater detail. We start with the learning setting before diving into the details regarding inference.

\noindent{\bf Learning:}
As mentioned before, when using a variational autoencoder, choosing appropriate $q$ and $p$ distributions is of crucial importance.  We show a high-level overview of our method in \figref{fig:HighLevel} and choose LSTM models for the encoder ($q$-distribution) and decoder ($p$-distribution). Learning amounts to  finding the parameters $\phi$ and $\theta$ of both modules. %, \ie, the word embeddings, the embedding for the LSTM units and the output mappings.
We detail our choice for both distributions in the following and provide more information regarding the trainable parameters of the model.

{{\bf Q-distribution:}} 
The $q$-distribution encodes a given sentence and a given image signal into a latent representation. % into the mean and variance of a  Gaussian distribution defined on the encoding space. 
Since this embedding is only used during training we can assume images and questions to be available in the following.
Our technique to encode images and questions is based on long short-term memory (LSTM) networks~\cite{HochreiterNC1997}. We visualize the computations in \figref{fig:QDist}.

Formally,  we  compute an $F$-dimensional feature  $f(I)\in\mathbb{R}^F$ of the provided image $I$ using a neural net, \eg, the VGG net discussed by Simonyan and Zisserman~\cite{Simonyan14c}. The LSTM unit first maps the image feature linearly into its $H$ dimensional latent space using  a matrix $W_I\in\mathbb{R}^{H\times F}$. For simplicity we neglect bias terms here and in the following.

Moreover,  each $V$-dimensional 1-hot encoding $x_i\in x=(x_1, \ldots, x_T)$ selects an $E$-dimensional word embedding vector from the matrix $W_e\in\mathbb{R}^{E\times V}$, which is learned. %\ziyu{There is no weights or biases between $x_i$ and $W_e$. The 1-hot encoding $x_i$ is used purely to select 1 column ($E$ x 1) from $W_e$.} To this end, we learn an embedding matrix $W_{e}\in\mathbb{R}^{E\times V}$ \sout{and corresponding bias terms}. 
The LSTM unit employs another linear transformation using the matrix $W_{e,2}\in\mathbb{R}^{H\times E}$ to project the word embedding into the $H$ dimensional space used inside the LSTM cells. % which we ommit in this paper for notational convenience. \alex{this is not clear to me yet} %\alex{are there bias terms?} \unnat{Yes there are.} %\ziyu{Similarly, for every single 1-hot encoded word $x_i$ in $x = (x_1, \ldots, x_T)$, we embed it as a $W$-dimensional vector obtained from the multiplication of its 1-hot encoding and the word embeddings matrix $W_e$.} 
We leave usage of more complex embeddings such as~\cite{WangCVPR2016,GongECCV2014} to future work. 

Given the $F$-dimensional image feature $f(I)$ and the $E$-dimensional word embeddings, the LSTM  internally maintains an $H$-dimensional representation. We found that providing the image embedding in the first step and each word embedding in subsequent steps to perform  best. After having parsed the image embedding and the word embeddings, we extract the final hidden representation $h_T\in\mathbb{R}^H$ from the last LSTM step. We subsequently apply two linear transformations  to the final hidden representation in order to obtain the mean $\mu = W_\mu h_T$ and the log variance $\log(\sigma^2) = W_\sigma h_T$ of an $M$-variate Gaussian distribution, \ie, $W_\mu\in\mathbb{R}^{M\times H}$ and $W_\sigma\in\mathbb{R}^{M\times H}$.
During training a zero mean and unit variance is encouraged, \ie, we use the prior $p_\theta(z) = \cN(0,1)$ in \equref{eq:ProgramSimple}.

{{\bf P-distribution:}}
The $p$-distribution is used to reconstruct a question $\hat x$ given, in our case, the image representation $f(I)\in\mathbb{R}^F$, and an $M$-variate random sample $z$. During inference the sample is drawn from a standard normal $\cN(0,1)$.  During training, this sample is shifted and scaled by the mean $\mu$ and the variance $\sigma^2$ obtained as output  from the encoder (the reparameterization trick). For  the $p$-distribution and the $q$-distribution, we use the same image features $f(I)$, but learn a different word embedding matrix, \ie, for the decoder $W_d\in\mathbb{R}^{E\times V}$. We observe different embedding matrices for the encoder and decoder to yield better empirical results. Again we omit the bias terms. %\alex{are there bias terms?} \unnat{Yes there are}

%\unnat{we use different transformation matrices to convert the one-hot encoded words (length $V$) to word embeddings (length $W$). \sout{we use the same image and word embedding.} \alex{is this true?}}

\begin{table}[t]
\vspace{-0.5cm}
\centering
\begin{tabular}{ccccc}
\hline
\cellcolor[HTML]{EFEFEF} & \multicolumn{2}{c}{\cellcolor[HTML]{EFEFEF}{\color[HTML]{333333} BLEU}} & \multicolumn{2}{c}{\cellcolor[HTML]{EFEFEF}{\color[HTML]{333333} METEOR}} \\ \cline{2-5} 
\multirow{-2}{*}{\cellcolor[HTML]{EFEFEF}Sampling} & Average & Oracle & Average & Oracle \\ \hline
N1, 100 & \textbf{0.356} & 0.393 & \textbf{0.199} & 0.219 \\
N1, 500 & 0.352 & 0.401 & 0.198 & 0.222 \\
U10, 100 & 0.328 & 0.488 & 0.190 & 0.275 \\
U10, 500 & 0.326 & 0.511 & 0.186 & 0.291 \\
U20, 100 & 0.316 & 0.544 & 0.183 & 0.312 \\
U20, 500 & 0.311 & \textbf{0.579} & 0.177 & \textbf{0.342} \\ \hline
\end{tabular}
\vspace{-0.4cm}
\caption{\textbf{Accuracy metrics:} Maximum (over the epochs) of average and oracle values of BLEU and METEOR metrics. Sampling the latent space by uniform distribution leads to better oracle scores. Sampling the latent space by a normal distribution leads to better average metrics. Interpretation in \secref{subsec:evaluation}. Table for VQG-Flickr and VQG-Bing are similar and are included in the supplementary material.}
\label{tab:vqq_coco_accuracy}
\vspace{-0.3cm}
\end{table}

\begin{figure}[t]
\vspace{-0.5cm}
    \centering
    \setlength{\tabcolsep}{0pt}
    \def\arraystretch{0.8}
    \begin{tabular}{c}
    \includegraphics[width=0.45\textwidth]{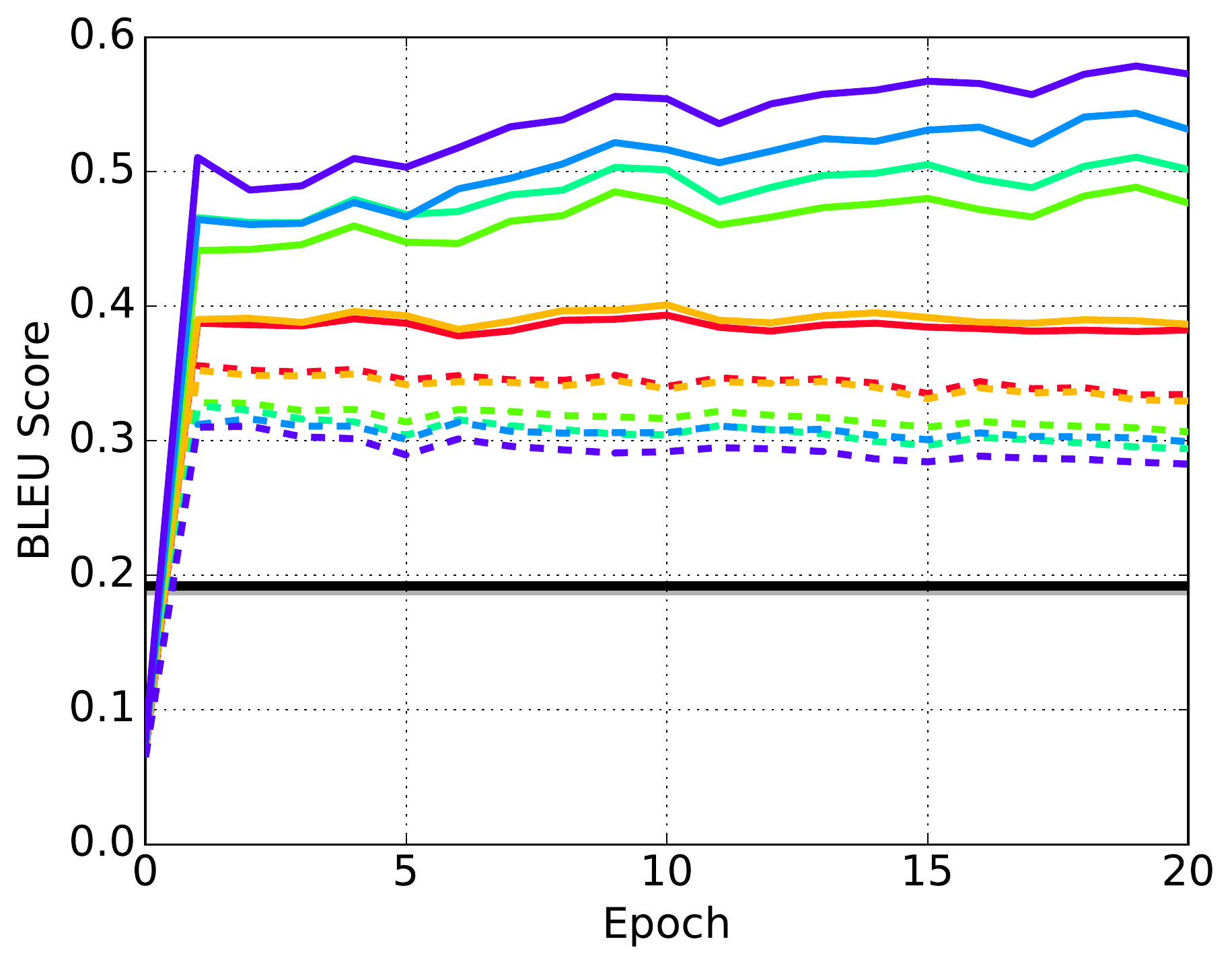}\\
    (a) Average-BLEU and oracle-BLEU score\\(Same legend as below)\\
    \includegraphics[width=0.45\textwidth]{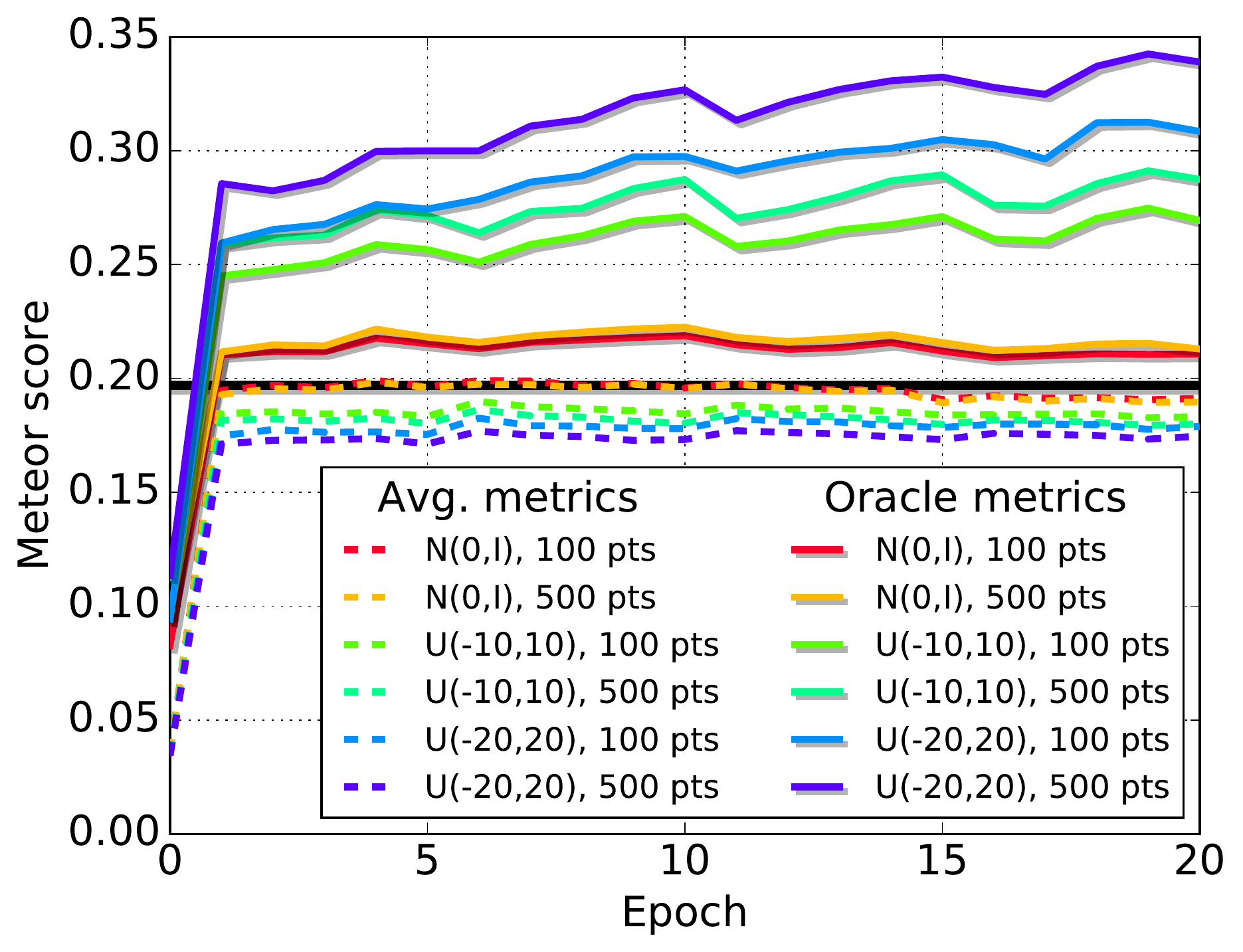}\\
    (b) Average-METEOR and oracle-METEOR scores\\
    \end{tabular}
    \vspace{-0.1cm}
%     \begin{subfigure}[b]{0.31\textwidth}
%         \centering
%         \includegraphics[width=\textwidth]{figures/vqg-coco__meteorvqa+coco+flickr+bing.pdf}
%         \caption{VQG-COCO}   
%         \label{fig:avg_uq_coco}
%     \end{subfigure}
%     \hfill
%     \begin{subfigure}[b]{0.31\textwidth}
%         \centering
%         \includegraphics[width=\textwidth]{figures/vqg-flickr__meteorvqa+coco+flickr+bing.pdf}
%         \caption{VQG-Flickr}
%         \label{fig:avg_uq_flickr}
%     \end{subfigure}
%     \hfill
%     \begin{subfigure}[b]{0.31\textwidth}
%         \centering
%         \includegraphics[width=\textwidth]{figures/vqg-bing__meteorvqa+coco+flickr+bing.pdf}
%         \caption{VQG-Bing}
%         \label{fig:avg_uq_bing}
%     \end{subfigure}
%     \vspace{-0.2cm}
\vspace{-0.2cm}
    \caption{\textbf{Accuracy metrics:} BLEU and METEOR scores for VQG-COCO. Experiments with various sampling procedures and results compared to the performance of the baseline model~\cite{VQG} (line in \textbf{black} color). VQG-Flickr and VQG-Bing results are similar and have been included in the supplementary material.}
    \label{fig:vqq_coco_accuracy}
     \vspace{-0.5cm}
\end{figure}

% \begin{table}[h]
% \centering
% \begin{tabular}{l c c c c }
% \hline
% Sampling & Bleu & Oracle Bleu & Meteor  & Oracle Meteor \\ \hline
% N1, 100 & \textbf{0.356} & 0.393 & \textbf{0.199 } & 0.219  \\
% N1, 500  &  0.352 &0.401       &0.198   &0.222      \\
% U10, 100 &0.328 & 0.488&0.19 & 0.275\\
% U10, 500 & 0.326&0.511 &0.186 & 0.291\\
% U20, 100 & 0.316& 0.544& 0.183 &0.312\\
% U20, 500 & 0.311& \textbf{0.579}& 0.177& \textbf{0.342}\\
% \hline
% \end{tabular}
% \centering
% \caption{\textbf{Accuracy metrics:} Maximum (over the epochs) ff average and oracle values of BLEU and METEOR metrics. Sampling the latent space by uniform distribution leads to a better oracle metrics. Interpretation in sec.~\ref{subsec:evaluation}.}
% \label{tab:vqq_coco_accuracy}
% \end{table}

Analogously to the encoder we use an LSTM network for decoding, which is visualized in \figref{fig:PDist}. Again we provide the $F$-dimensional image representation $f(I)$ as the first input signal. Different from the encoder we then provide as the input to the second LSTM unit a randomly drawn $M$-variate sample $z\sim\cN(0,1)$, which is shifted and scaled by the mean $\mu$ and the variance $\sigma^2$ during training. %as the input to the second LSTM unit. %\alex{Aren't we transforming the random sample via a linear matrix to fit the embedding dimension? Otherwise the dimensions don't match, do they? Isn't that also missing in Fig. 5?}
Input to the third and all subsequent LSTM units is an $E$-dimensional embedding of the start symbol and subsequently the word embeddings $W_d x_i$. As for the encoder, those inputs are transformed by the LSTM units into its $H$-dimensional operating space.

To compute the output we use the $H$-dimensional hidden representation $h_i$ which we linearly transform via a $V\times H$-dimensional matrix into the $V$-dimensional vocabulary vector of logits, on top of which a softmax function is applied. This results in a probability distribution $p_0$ over the vocabulary at the third LSTM unit. During training, we maximize the predicted log-probability of the next word in the sentence, \ie, $x_1$. Similarly for all subsequent LSTM units.%} \sout{and from which we pick the output space \sout{from  which we decode the first word $x_0$ using the maximizing entry.}}

In our framework, we jointly learn the word-embedding $W_e\in\mathbb{R}^{E\times V}$ together with the $V\times H$-dimensional output embedding, the $M\times H$-dimensional encoding, and the LSTM projections to the $H$-dimensional operating space. The number of parameters (including the bias terms) in our case are $2VE$ from the word embeddings matrix, one for the encoder and another for the decoder; $HV+V$ as well as $2(HM+M)$ from the output embedding of the decoder and the encoder respectively;  $(FH+H) + 2(EH+H) + (MH+H) + (HH+H)$ internal LSTM unit variables. %Importantly, the input embedding is identical for the encoder ($p$-distribution) and the decoder ($q$-distribution).

%\ziyu{parameters:

%In encoder, weights and biases from image to lstm (omitted in figure), word embeddings matrix, weights and biases from word embedding to lstm (omitted in figure), lstm parameters, weights and biases from lstm's hidden to $\mu$ and $\sigma^2$.

%In decoder, weights and biases from image feature to lstm (omitted in figure), weights and biases from latent to lstm (omitted in figure), word embeddings matrix, weights and biases from word embedding to lstm (omitted in figure), lstm parameters, weights and biases from lstm's hidden to vocabulary (p).}

\noindent{\bf Inference:}
After having learned the parameters of our model on a dataset consisting of pairs of images and questions we obtain a decoder that is able to generate questions given an embedding $f(I)\in\mathbb{R}^F$ of an image $I$ and a randomly drawn $M$-dimensional sample $z$ either from a standard normal or a uniform distribution. %\ziyu{(Not necessarily normal distribution)}. \alex{we'll mention that in the experimental setting}.
Importantly for every different choice of input vector $z$ we  generate a new question $x = (x_1, \ldots, x_T)$.

Since no groundtruth $V$-dimensional embedding is available, during inference, we use the prediction from the previous timestep as the input to predict the word for the current timestep. 

%\ziyu{During inference, the only difference from training is that whenever we obtain a distribution over the vocabulary at a given LSTM step, we pick the word with highest probability and use it as the input into the next LSTM step. The process keeps going until the end symbol is the chosen word or a maximum number of iterations is reached.}

%\unnat{I think this is the case for only test time:} This generated word $x_{t-1}$ is then embedded into an $N$-dimensional vector space before being provided  as the input to the following LSTM unit which generates the $x_t$-th word. The process continues until the termination symbol is generated.

\subsection{Implementation details}
Throughout, we used the  4096-dimensional \textit{fc6} layer of the 16-layer VGG model~\cite{Simonyan14c} as our image feature $f(I)$, \ie, $F = 4096$. We also fixed the 1-hot encoding of the vocabulary, \ie, $V = 10849$, to be the number of words we collect from our datasets (VQA+VQG, detailed in the next section). 
We investigated  different dimensions for the word embedding ($E$),  the hidden representation ($H$), and the encoding space ($M$). We found $M=20$, $H=512$, and $E=512$ to provide enough representational power for training on roughly $400,000$ questions obtained from roughly $126,000$ images.

We found an initial learning rate of $0.01$ for the first $5$ epochs to reduce the loss quickly and to give good results. We reduce this learning rate by half every 5 epochs. % \unnat{A good reason for this here?}.

\begin{table}[t]
\vspace{-0.5cm}
\centering
\begin{tabular}{cclcl}
\hline
\cellcolor[HTML]{EFEFEF} & \multicolumn{2}{c}{\cellcolor[HTML]{EFEFEF}{\color[HTML]{333333} }} & \multicolumn{2}{c}{\cellcolor[HTML]{EFEFEF}{\color[HTML]{333333} }} \\
\multirow{-2}{*}{\cellcolor[HTML]{EFEFEF}Sampling} & \multicolumn{2}{c}{\multirow{-2}{*}{\cellcolor[HTML]{EFEFEF}{\color[HTML]{333333} \begin{tabular}[c]{@{}c@{}}Generative Strength\\ (\%)\end{tabular}}}} & \multicolumn{2}{c}{\multirow{-2}{*}{\cellcolor[HTML]{EFEFEF}{\color[HTML]{333333} \begin{tabular}[c]{@{}c@{}}Inventiveness\\ (\%)\end{tabular}}}} \\ \hline
N1, 100 & \multicolumn{2}{c}{1.98} & \multicolumn{2}{c}{10.76} \\
N1, 500 & \multicolumn{2}{c}{2.32} & \multicolumn{2}{c}{12.19} \\
U10, 100 & \multicolumn{2}{c}{9.82} & \multicolumn{2}{c}{18.78} \\
U10, 500 & \multicolumn{2}{c}{16.14} & \multicolumn{2}{c}{24.32} \\
U20, 100 & \multicolumn{2}{c}{22.01} & \multicolumn{2}{c}{19.75} \\
U20, 500 & \multicolumn{2}{c}{\textbf{46.10}} & \multicolumn{2}{c}{\textbf{27.88}} \\ \hline
\end{tabular}
\vspace{-0.3cm}
\caption{\textbf{Diversity metrics:} Maximum (over the epochs) value of generative strength and inventiveness on the VQG-COCO test set. Sampling the latent space by a uniform distribution leads to more unique questions as well as more unseen questions. Table for VQG-Flickr and VQG-Bing are similar and are included in the supplementary material.}
\label{tab:vqq_coco_diversity}
\vspace{-0.3cm}
\end{table}

\begin{figure}[t]
\vspace{-0.5cm}
\setlength{\tabcolsep}{0pt}
    \def\arraystretch{0.8}
  \centering
    \begin{tabular}{c}
    \includegraphics[width=0.45\textwidth]{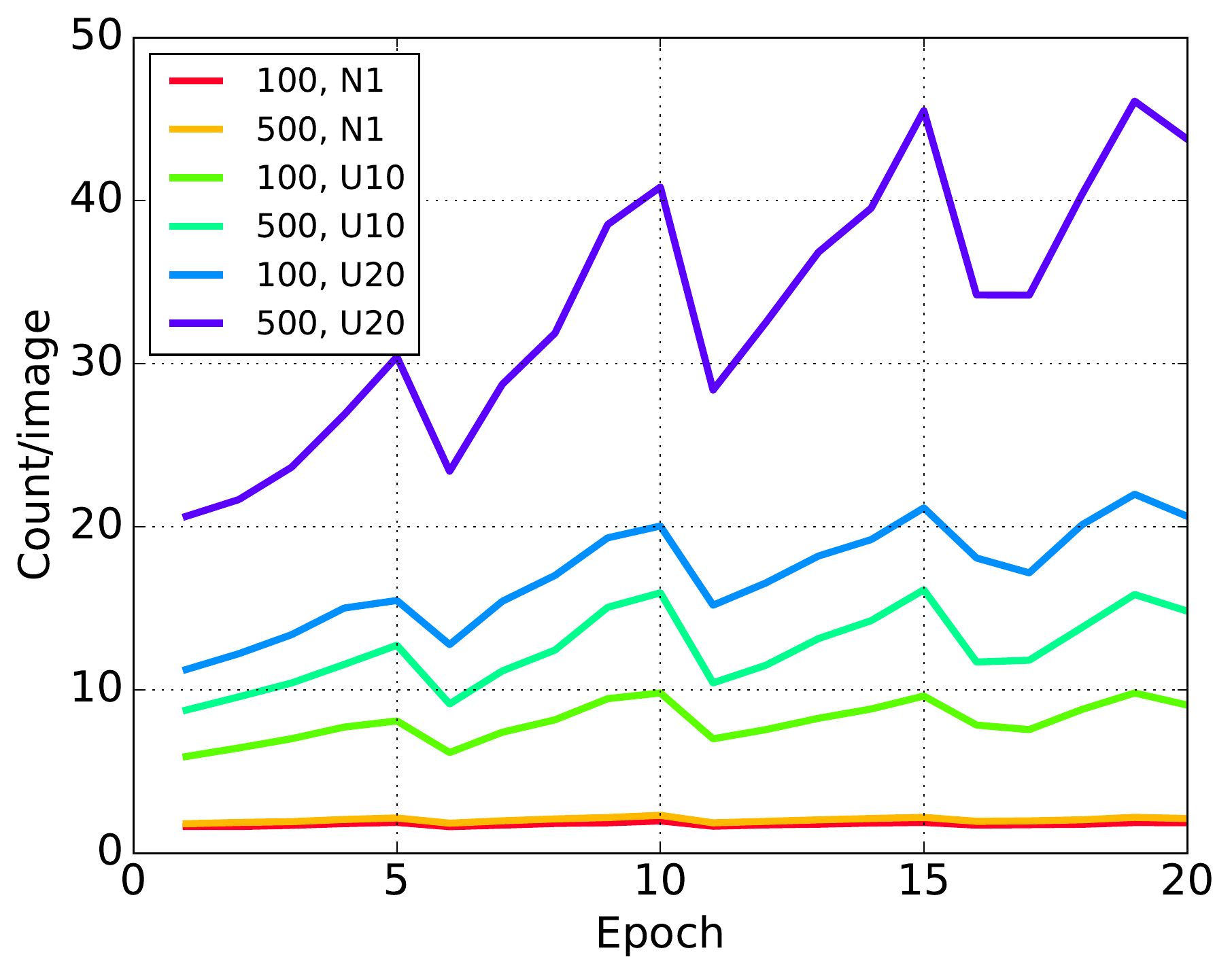}\\
    (a) \textbf{Generative strength:} Number of unique\\questions averaged over the number of images.\\
    \includegraphics[width=0.45\textwidth]{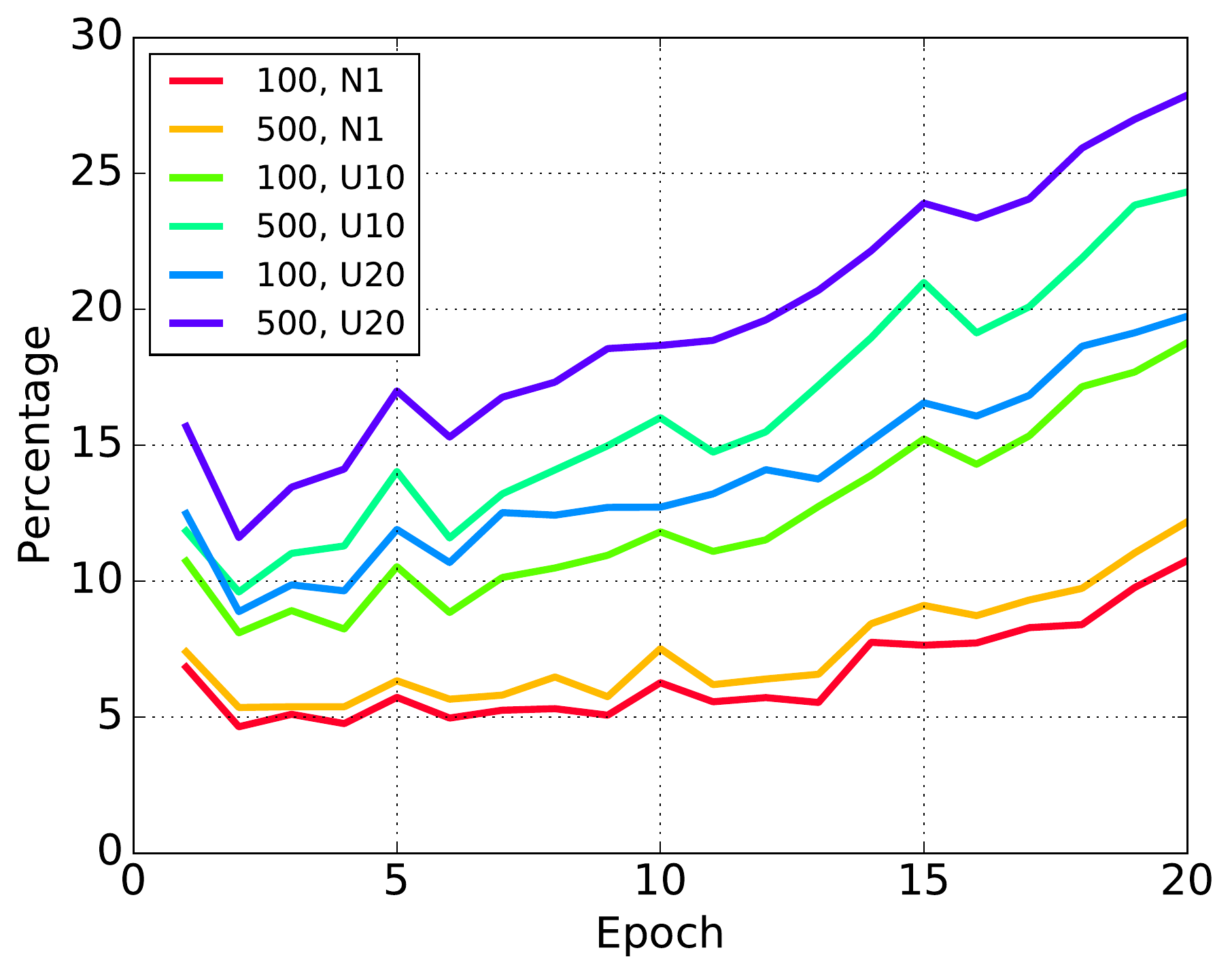}\\
    (b) \textbf{Inventiveness:} $\frac{\text{Unique questions which were never seen in training set}}{\text{Total unique questions for that image}}$\\
    \end{tabular}
    \vspace{-0.1cm}
    \caption{\textbf{Diversity metrics:} Generative strength and Inventiveness, averaged over all the images in the VQG-COCO test set. VQG-Flickr and VQG-Bing results are similar and are included in  the supplementary material.}
    \label{fig:vqq_coco_diversity}
    \vspace{-0.2cm}
\end{figure}

%% file: exp.tex
\section{Experiments}

In the following we evaluate our proposed technique on the  VQG dataset~\cite{VQG} and present a variety of different metrics to demonstrate the performance. We first describe the datasets and metrics, before providing our results.

\subsection{Datasets:}

\begin{figure*}
\vspace{-0.5cm}
    \centering
    \begin{subfigure}[b]{0.30\textwidth}
        \centering
        \includegraphics[width=\textwidth]{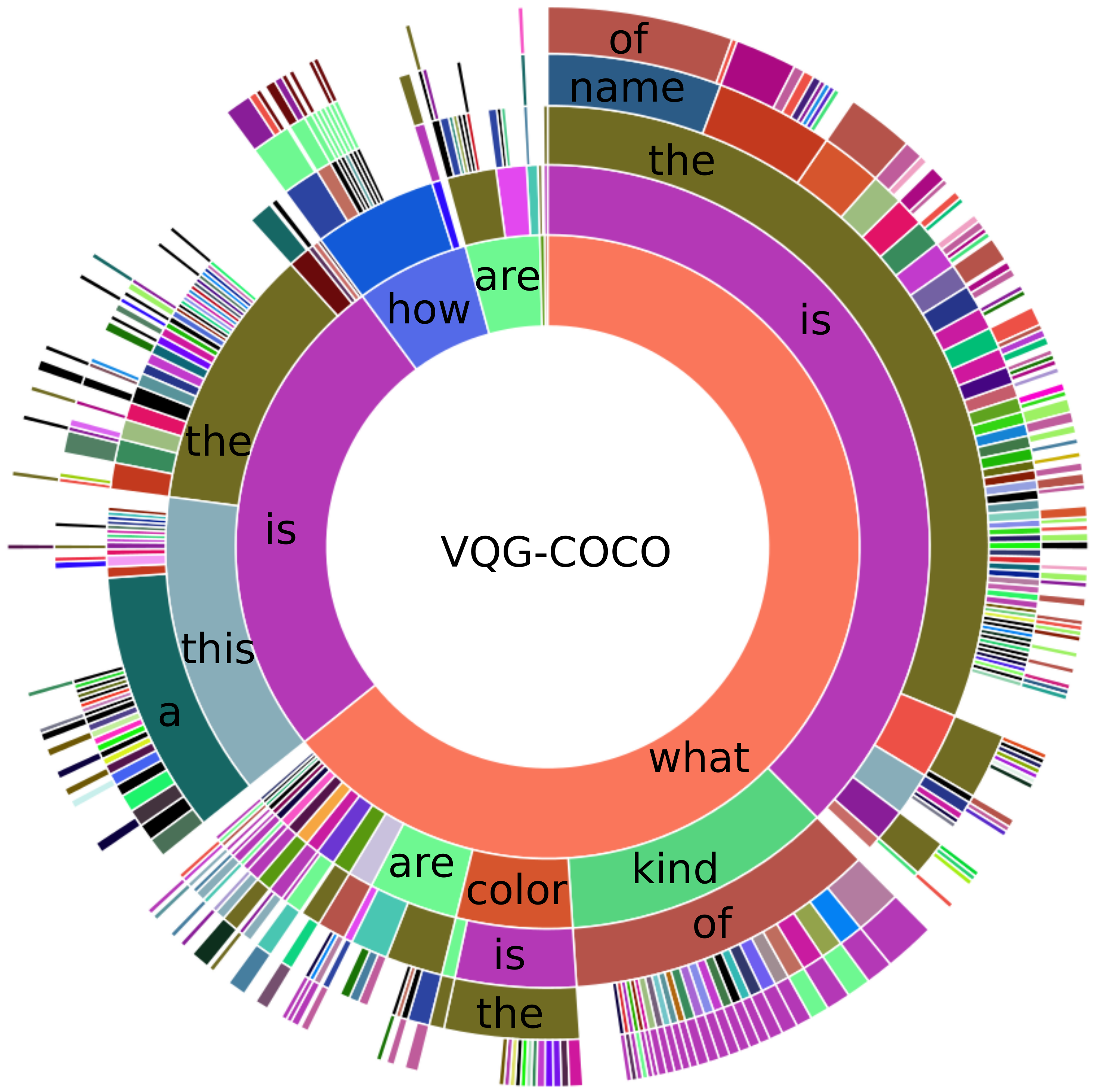}
        \caption{VQG-COCO}   
        \label{fig:sunburst_coco}
    \end{subfigure}
    \hfill
    \begin{subfigure}[b]{0.30\textwidth}
        \centering
        \includegraphics[width=\textwidth]{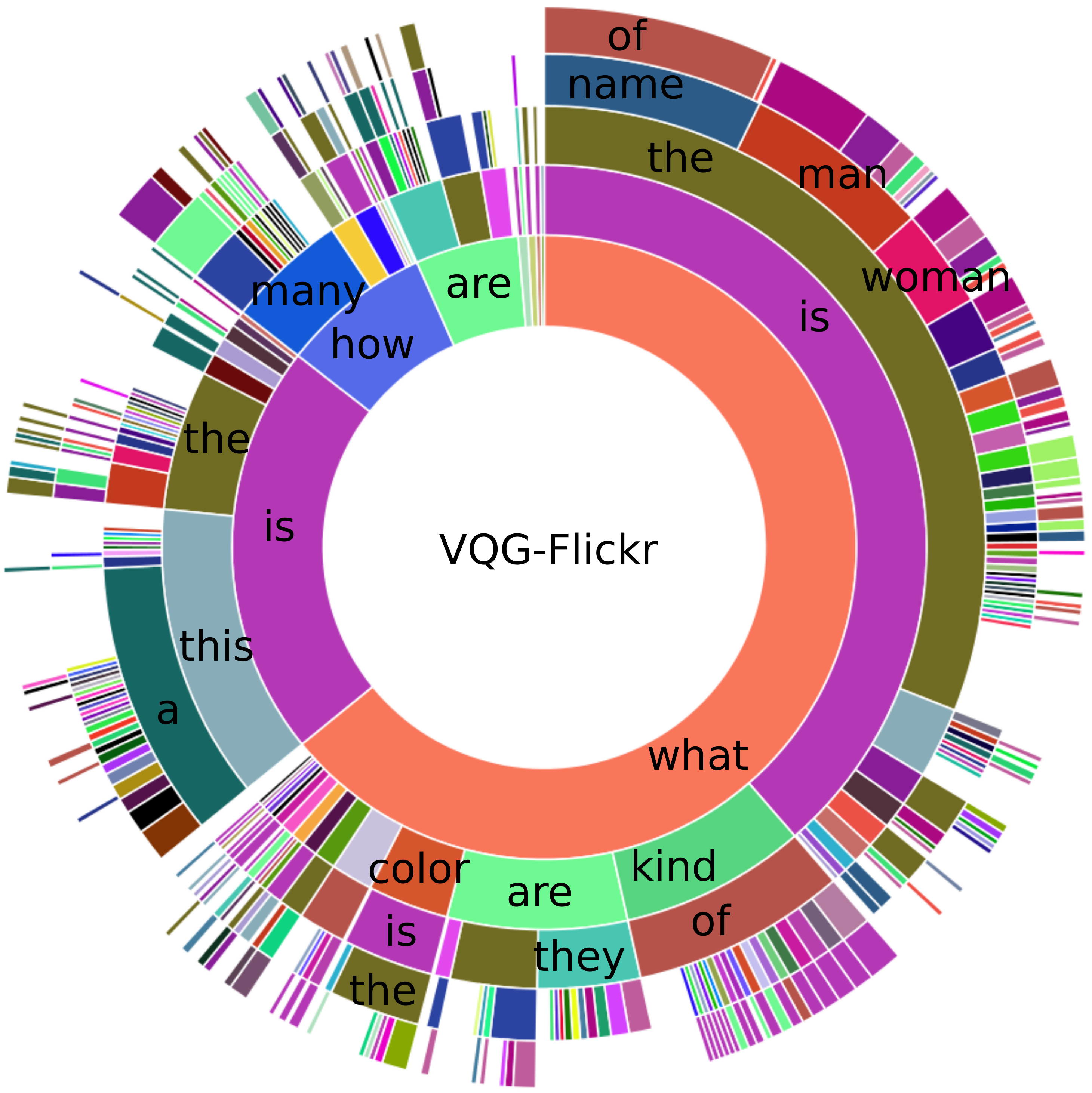}
        \caption{VQG-Flickr}
        \label{fig:sunburst_flickr}
    \end{subfigure}
    \hfill
    \begin{subfigure}[b]{0.30\textwidth}
        \centering
        \includegraphics[width=\textwidth]{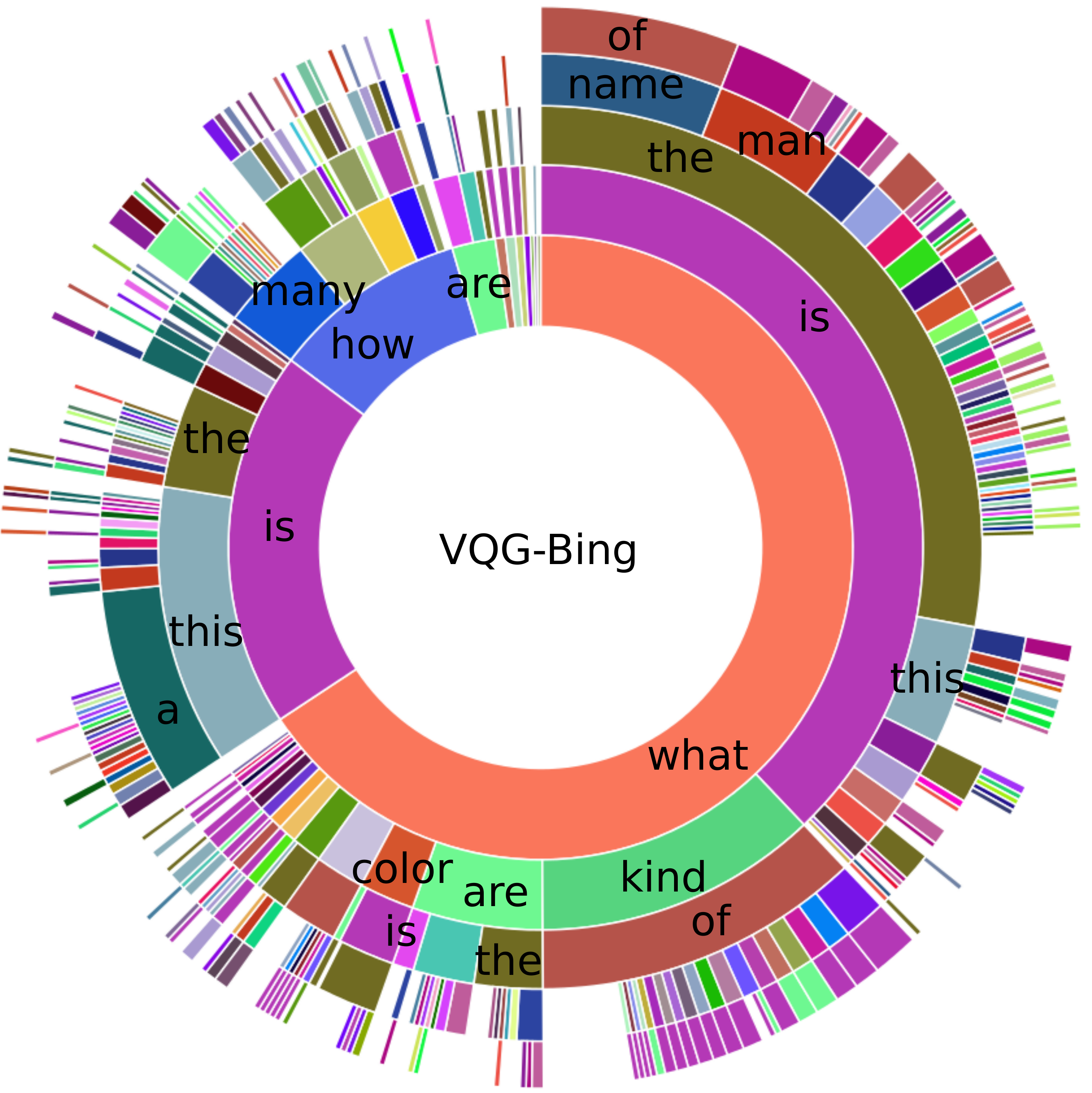}
        \caption{VQG-Bing}
        \label{fig:sunburst_bing}
    \end{subfigure}
    \vspace{-0.4cm}
    \caption{\textbf{Sunburst plots for diversity}: Visualizing the diversity of questions generated for each of VQG datasets. The $i^{\text{th}}$ ring captures the frequency distribution over words for the $i^{\text{th}}$ word of the generated question. The angle subtended at the center is proportional to the frequency of the word. While some words have high frequency, the outer rings illustrate a fine blend of words similar to the released dataset~\cite{VQG}. We restrict the plot to 5 rings for easy readability.}
    \label{fig:sunburst}
    \vspace{-0.5cm}
\end{figure*}

\noindent{\bf VQA dataset:} The images of the VQA dataset~\cite{AnatolICCV2015} are obtained from the MS COCO dataset~\cite{lin2014microsoft}, and divided into $82,783$ training images, $40,504$ validation images and $40,775$ testing images. Each image in the training and validation sets is annotated with 3 questions. The answers provided in the VQA dataset are not important for the problem we address.

\noindent{\bf VQG datasets:} The Visual Question Generation~\cite{VQG} dataset consist of images from MS COCO, Flickr and Bing. Each of these sets consists of roughly $5,000$ images and $5$ questions per image (with some exceptions). Each set is split into $50\%$ training, $25\%$ validation and $25\%$ test. VQG is a dataset of natural and engaging questions, which goes beyond simple literal description based questions.

% \begin{figure}[t]
% \centering
% \includegraphics[width=\linewidth]{figures/vqa+coco_coco_test_U20_100_all_questions.png}
% \caption{Generative diversity represented by sunburst images of the \textbf{VQG-COCO} test set. (Sampling 100 latent space points per image, from the U(20) distribution)\alex{is it possible to get rid of the highlighting in those plots? just consumes space}}
% \label{fig:sunburst_coco}
% \end{figure}

% \begin{figure}[t]
% \centering
% \includegraphics[width=\linewidth]{figures/20-100pts_flickr_test_all_questions.png}
% \caption{Generative diversity represented by sunburst images of the \textbf{VQG-Flickr} test set. (Sampling 100 latent space points per image, from the U(20) distribution)\alex{is it possible to get rid of the highlighting in those plots? just consumes space}}
% \label{fig:sunburst_flickr}
% \end{figure}

The VQG dataset targets the ambitious %\unnat{Do we want to call it ambiguous?} 
problem of `natural question generation.' However, due to its very small size, training of larger scale generative models that fit the high-dimensional nature of the problem is a challenge. Throughout our endeavor we found a question dataset size similar to the size of the VQA dataset to be extremely beneficial.

\noindent{\bf VQA+VQG dataset} 
To address this issue, we combined the VQA and VQG datasets. VQA's sheer size provides enough data to learn the parameters of our LSTM based VAE model. Moreover, VQG adds additional diversity due to the fact that questions are more engaging and natural. 
The combined training set has $125,697$ images (VQA training + VQA validation + VQG-COCO training - VQG-COCO validation - VQG-COCO test + VQG-Flickr training + VQG-Bing training) and a total of $399,418$ questions. We ensured that there is absolutely no overlap between the images we train on and the images we evaluate.  Since different images may have the same question, the number of unique questions out of \textit{all} training question is $238,699$.
%This training set has ... (248986 / 244078) images, a total of ... (761785 / 737228) question. Two different images might have the same question, hence we also found the unique questions out of \textit{all} the training question. Unique training questions are ... (237420 / 220409).
%}
%}
% \paragraph{y dataset:}
% \paragraph{z dataset:}

\subsection{Metrics}

\noindent{\bf BLEU:} BLEU, originally designed for evaluating the task of machine translation, was one of the first metrics that achieved good correlation with human judgment. It calculates `modified' $n$-gram precision and combines them to output a score between 0 to 1. BLEU-4 considers up to 4-grams and has been used widely for evaluation of existing works on machine translation, generating captions and questions.

% it machine-translation origin implies that the whole idea behind it is to evaluate how close a hypothesis is to references (in terms of \unnat{`modified'} $n$-gram precision). However, in our case where the goal is to generate questions as diverse as possible, there is no prior expectation of what content in the image should be asked about, as opposed to the task of machine translation where there is a clear target. Thus, comparing our generated questions against manually annotated questions may be a sub-optimal way for evaluating the quality of the generated questions.
%score is based on n-gram precision. This implies that even if a hypothesis is a good question for the image by human judgment, the score may still suffer if the words in the hypothesis have little overlap with reference questions.
% Why good and why bad}

\noindent{\bf METEOR:} The METEOR score is another machine translation metric which correlates well with human judgment. An F-measure is computed based on word matches. The best among the scores obtained by comparing the candidate question to each reference question is returned. In our case there are five reference questions for each image in VQG test sets. Despite BLEU and METEOR having considerable shortcomings (details in~\cite{vedantam2015cider}), both are popular metrics of comparison.

\noindent{\bf Oracle-metrics:} There is a major issue in directly using machine translation metrics such as BLEU and METEOR for evaluating generative approaches for caption and question generation. Unlike other approaches which aim to create a caption or question which is similar to the `reference,'  generative methods like~\cite{VinyalsCVPR2015,VijayakumarARXIV2016} and ours  produce multiple diverse and creative results which might not be present in the dataset. Generating a dataset which contains all possible questions is desirable but illusive. %This is a side-effect of the fact that the references provided in datasets cannot span the whole ground truth, as there are just too many different questions/captions that one could generate based on a sufficiently complex scenery.
Importantly, our algorithm may not necessarily generate questions which are only simple variations of a groundtruth question as sampling of the latent space provides the ability to produce a wide variety of questions. 
%Specifically, our model samples the latent space and has the ability to create questions  which are not provided as groundtruth. It may also create questions which are simple variations (more like machine translations) of the reference questions provided in the datasets.
\cite{VinyalsCVPR2015,VijayakumarARXIV2016} highlight this very issue, and combat it by stating their results using what~\cite{VijayakumarARXIV2016} calls \textit{oracle-metrics}. Oracle-BLEU, for example, is the maximum value of the BLEU score over a list of $k$ potential candidate questions. %\unnat{It signifies the best performance of a generative model on a particular metric.}
Using these metrics we compare our results to approaches such as~\cite{VQG} which infer one question per image aimed to be similar to the reference question.

% However, it has the same problem as BLEU, because METEOR also relies on reference questions to evaluate the quality of a hypothesis. After all, the difficulty of evaluating generative models lies in the absence of ground truth.} Why good and why bad

% \paragraph{z:} Why good and why bad

\noindent{\bf Diversity score:} Popular machine translation metrics such as BLEU and METEOR provide an insight into the accuracy of the generated questions. In addition to showing that we perform well on these metrics, we felt a void for a metric which captures the \textit{diversity}. This metric is particularly important when being interested in an engaging system. To demonstrate diversity, we evaluate our model on two intuitive metrics which could serve as relevant scores for future work attempting to generate diverse questions. The two metrics we use are average number of unique questions generated per image, and the percentage among these questions which have never been seen at training time. The first metric  assesses what we call the \textit{generative strength} and the latter represents the \textit{inventiveness} of models such as ours.

\subsection{Evaluation}
\label{subsec:evaluation}
In the following we first evaluate our proposed approach quantitatively  using the  aforementioned metrics, \ie, BLEU score, METEOR score and the proposed diversity score.
Subsequently, we provide additional qualitative results illustrating the diversity of our approach. We show results for two sampling techniques, \ie, sampling $z$ uniformly and sampling  $z$ using a normal distribution.

\noindent{\bf BLEU:} BLEU score approximates human judgment at a \textit{corpus level} and does not necessarily correlate well if used to evaluate sentences individually. Hence we state our results for the corpus-BLEU score (similar to~\cite{VQG}). The best performing models presented in~\cite{VQG} have corpus-BLEU of $0.192$, $0.117$ and $0.123$ for VQG-COCO, VQG-Flickr and VQG-Bing datasets respectively. To illustrate this baseline, we highlight these numbers using black lines on our plots in  \figref{fig:vqq_coco_accuracy}~(a). % to set the baseline of comparison.
% Our model's oracle-BLEU scores are much higher with the $U(-20,20)$ sampling scheme using 500 points.

% Camera ready edit: vv
% giving an oracle-BLEU of 0.5188 (COCO), 0.4829 (Flickr), and 0.4821 (Bing) when averaged over 21 training epochs, and a maximum value of  0.5786 (COCO), 0.5414 (Flickr), and 0.5382 (Bing). The average-BLEU scores averaged over 21 training epochs are  0.2826 (COCO),  0.2672 (Flickr), and  0.2607 (Bing), and the maximum average-BLEU scores over 21 epochs are  0.3109 (COCO),  0.2987 (Flickr), and  0.2912 (Bing).
% We also observe that the oracle-BLEU score has an increasing trend with the epochs. 
% Camera ready edit: ^^
%\sout{We also observe that the Bleu score increases initially but overfitting may cause it to drop during later stages of the training.}

\noindent{\bf METEOR:} In \figref{fig:vqq_coco_accuracy}~(b) we illustrate the METEOR score for our model on the VQG-COCO dataset. Similar to BLEU, we compute corpus-level scores as they have much higher correlation with human judgment. The best performing models presented in~\cite{VQG} have corpus-METEOR of $0.197$, $0.149$ and $0.162$ for VQG-COCO, VQG-Flickr and VQG-Bing datasets respectively. To illustrate this baseline, we highlight these numbers using black lines on our plots in  \figref{fig:vqq_coco_accuracy}~(b). 
In \tabref{tab:vqq_coco_accuracy} we compile the corpus and oracle metrics for six different sampling schemes. The sampling for results listed towards the bottom of the table is less confined. %We relax the prior of the latent space (increased exploration) as we go down in the table. 
The closer the sampling scheme is to the $\cN(0,1)$, the closer is our generated corpus of questions to the reference question  of the dataset. On the other hand, the more  exploratory the sampling scheme, the better is the best candidate (hence, increasing oracle metrics).

\noindent{\bf Diversity:}  \figref{fig:vqq_coco_diversity} illustrates the \textit{generative strength} and \textit{inventiveness} of our model with different sampling schemes for $z$. For the best $z$ sampling mechanism of $U(-20,20)$ using 500 points, we obtained on average 46.10 unique questions per image (of which 26.99\% unseen in the training set) for COCO after epoch 19;  For Flickr, 59.57 unique questions on average (32.80\% unseen) after epoch 19; For Bing, 63.83 unique questions on average (36.92\% unseen) after epoch 15. %The corresponding average `unseen' question percentage is ... and maximum is \ziyu{I think reporting the max number of unique questions out of all epochs with its ratio of ``unseen'' questions is sufficient.}... .
In \tabref{tab:vqq_coco_diversity}, even though the training prior over the latent space is a $\cN(0,1)$ distribution, sampling from the exploratory $U\text{(-20,20)}$ distribution leads to better diversity of the generated questions.

\begin{figure}[t]
\vspace{-0.5cm}
\centering
\includegraphics[width=0.9\linewidth]{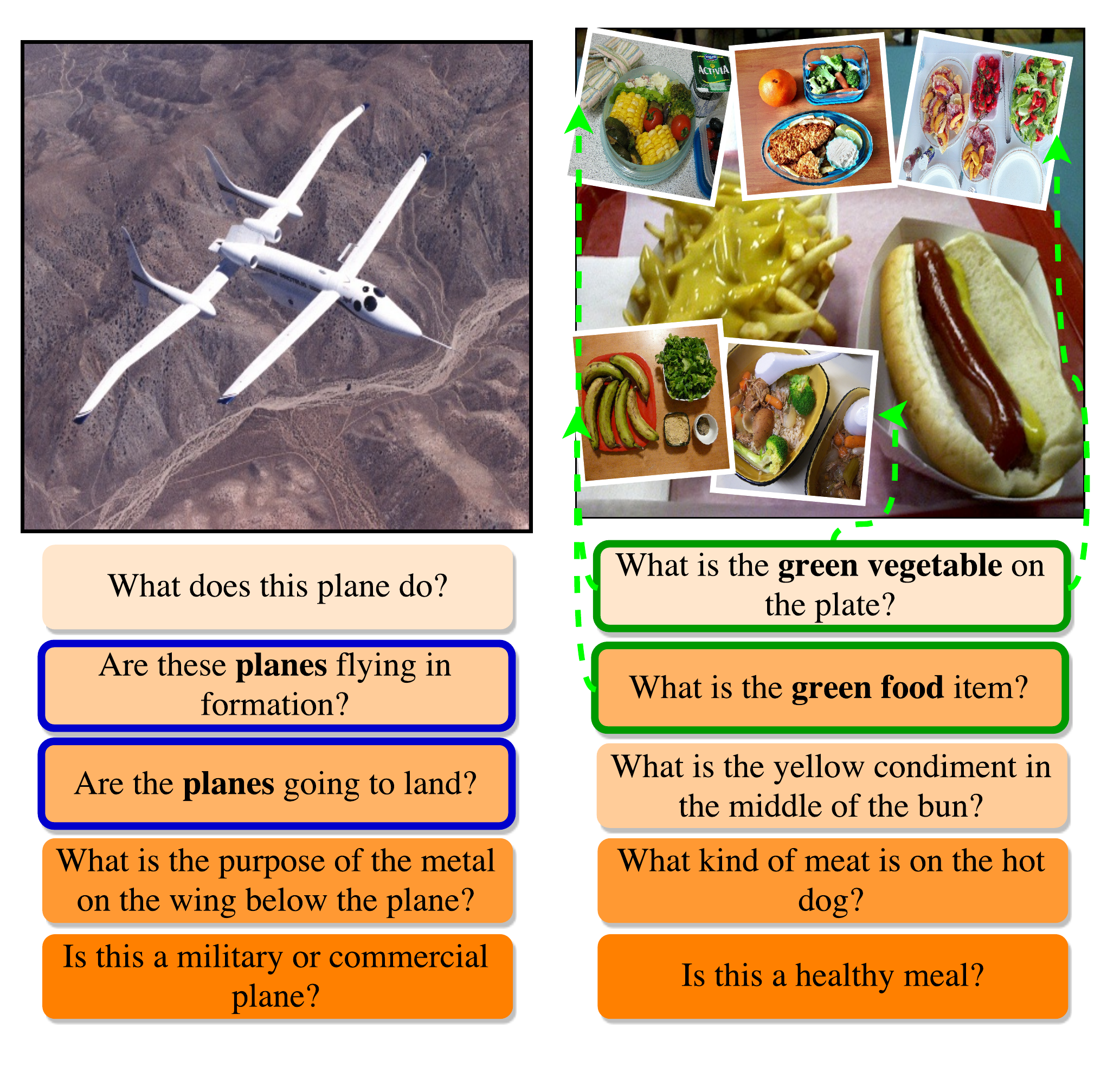}
\vspace{-0.5cm}
\caption{\textbf{\textit{Recognition} and \textit{co-occurrence} based failure cases:} \textit{Left:} A special aircraft is recognized as multiple `airplanes' (two sets of wings instead of one may cause the confusion), therefore, erroneous questions (marked in \textcolor{blue}{blue}) arise. \textit{Right:} Due to very frequent co-occurrence of green vegetable/food/fruit in food images, our VQG model generates questions (marked in \textcolor{green}{green}) about green vegetables even when they are missing. The five small images are few examples of how training set food images almost always contain greens.}
\label{fig:fail}
\vspace{-0.4cm}
\end{figure}

To further illustrate the diversity of the generated questions we use the \href{http://bl.ocks.org/mbostock/4063423}{sunburst} plots shown in \figref{fig:sunburst} for the COCO, Flickr and Bing datasets. Despite the fact that a large number of questions start with ``what'' and ``is,'' we still observe a quite reasonable amount of diversity.

\noindent{\bf Qualitative results:} In \figref{fig:qual} we show  success cases of our model. A range of literal to inferential questions are generated by our model, some requiring strong prior (human-like) understanding of objects and their interaction. In previous subsections  we showed that our model does well on metrics of accuracy and diversity. In \figref{fig:fail} we illustrate two categories of failure cases. \textit{Recognition failures}, where the pre-learned visual features are incapable of capturing correctly the information required to formulate diverse questions. As illustrated by the image of a complex aircraft which appears similar to two airplanes. Hence, our system generates questions coherent to such a perception.

Second are \textit{co-occurrence based failures}. This is illustrated using the image of fries and a hot dog. In addition to some correct questions, some questions on green food/fruit/vegetables inevitably pop up in food images (even for images without any greens). Similarly, questions about birds are generated in some non-bird images of trees. This could be accounted to very frequent co-occurrence of reference questions on greens or birds whenever an image contains food or trees, respectively.

%Training and end-to-end CNN + VAE system capable of learning features weights best adapted for the VQG task, would be a start for future work. We also observed incorrect natural language structure for a small percentage of questions. We believe a more powerful sentence generation model such as multiple layers of LSTMs could help to resolve this issue. In future work we plan to investigate all these cases more carefully and understand which portion of the image does our VQG model  attend to.

%% file: conc.tex
%!TEX root = egpaper_for_review.tex
\section{Conclusion}
In this paper we propose to combine the advantages of variational autoencoders with long-short-term-memory (LSTM) cells to obtain a ``creative'' framework that is able to generate a diverse set of questions given a single input image. We demonstrated the applicability of our framework on a diverse set of images and envision it being applicable in domains such as computational education, entertainment and for driving assistants \& chatbots. 
In the future we plan to use more structured reasoning~\cite{ChenSchwingICML2015, SchwingARXIV2015,LondonNIPS2016,MeshiNIPS2015,SchwingICML2014,SchwingNIPS2012}.